\documentclass[letterpaper, 10pt, conference]{ieeeconf}  

\IEEEoverridecommandlockouts                              

\usepackage{graphicx} 

\usepackage{amsmath,amssymb,amsfonts}
\usepackage{cite}
\usepackage{multirow}
\usepackage{algorithmic}
\usepackage[ruled,vlined,lined,linesnumbered,boxed,commentsnumbered]{algorithm2e}
\usepackage{booktabs}
\usepackage[caption=false]{subfig}
\usepackage{color}
\usepackage{balance}
\usepackage{footmisc}
\usepackage{fancyhdr}

\fancypagestyle{firstpage}{%
  \chead{This paper has been accepted for publication in the IEEE Int. Conf. Robot. Autom. (ICRA), 2020. @ IEEE}
}

\title{\LARGE \bf
Closed-Loop Benchmarking of Stereo Visual-Inertial SLAM Systems: Understanding the Impact of Drift and Latency on Tracking Accuracy
}

\author{Yipu Zhao$^{1}$, Justin S. Smith$^{1}$, Sambhu H. Karumanchi$^{1}$, 
  and Patricio A. Vela$^{1}$%
\thanks{$^{1}$%
Y. Zhao, J.S. Smith, S.H Karumanchi, and P.A. Vela  
are with the School of Electrical and Computer Engineering, 
Georgia Institute of Technology, Atlanta, Georgia, USA. 
{\tt\small \{yipu.zhao, jssmith, skarumanchi3, pvela\}@gatech.edu}. 
This work was supported by the National Science Foundation 
(Award \#1816138).
}%
}

\usepackage{url}
\usepackage{./slam_macros}

\graphicspath{ {./images/}}

\begin{document}

\maketitle
\thispagestyle{firstpage}

\begin{abstract}
Visual-inertial SLAM is essential for robot navigation in 
GPS-denied environments, e.g. indoor, underground. 
Conventionally, the performance of visual-inertial SLAM 
is evaluated with open-loop analysis, with a focus on 
the drift level of SLAM systems.  
In this paper, we raise the question on the importance of 
visual estimation latency in closed-loop navigation tasks, 
such as accurate trajectory tracking. 
To understand the impact of both drift and latency on visual-inertial
SLAM systems, a closed-loop benchmarking simulation is conducted, where 
a robot is commanded to follow a desired trajectory using the feedback 
from visual-inertial estimation.  
By extensively evaluating the trajectory tracking performance of 
representative state-of-the-art visual-inertial SLAM systems, 
we reveal the importance of latency reduction in visual estimation 
module of these systems.  The findings suggest directions of future 
improvements for visual-inertial SLAM.
\end{abstract}


%
\section{Introduction}

Vision-based state estimation techniques, such as Visual Odometry (VO)
and Visual Simultaneous Localization and Mapping (VSLAM), are essential
for robots to autonomously navigate through unmapped scenes.  
VO often forgets the sensed world structure, while VSLAM retains a
long-term map of the traversed world.  In the absence of absolute
position signals such as from GPS, VO/VSLAM complements traditional
wheel/inertial-based odometry.  

Compared with VO/VSLAM that relies on vision sensor only, visual-inertial 
SLAM (VI-SLAM) uses the two complementary data streams to achieve better
accuracy and robustness, and higher frequency, of state estimation.  
The visual sensor provides accurate, yet sparse and delayed measurements
of absolute landmarks in the environment.  
Estimation drift is mitigated by observing and matching landmarks with a
long but potentially intermittent measurement history. 
%
The inertial sensor provides high-rate, almost-instantaneous, yet
drifting measurements of robot motion.  Inertial measurements compensate
for short duration visual feature loss (e.g. in texture poor settings).  
The pose estimates of a VI-SLAM system can be sent to a controller as a 
high-quality feedback signal in support of trajectory tracking
as the mobile robot navigates through an environment.  

While the ultimate use case of VI-SLAM in robotics is closed-loop navigation, 
traditional benchmarking of VI-SLAM employs open-loop analysis, 
i.e., the SLAM output doesn't affect actual robot actuation and future sensory input. 
Though reflecting the estimation drift level of VI-SLAM, 
open-loop evaluation fails to fully address the coupled impact of navigation
and VI-SLAM estimation during closed-loop operation.
For targeted closed-loop navigation, it is hard to gain insights on 
VI-SLAM from published open-loop benchmark scores.  
To address this benchmarking gap, we present an open-source 
\cite{ZhEtAl_SLAM_CLBenchGit}, reproducible benchmarking simulation for
closed-loop VI-SLAM evaluation, and the outcomes from evaluating several
VI-SLAM methods using it.
Reproducible, closed-loop benchmarking should serve to guide future
VI-SLAM research for mobile robotics.


\begin{figure}[t]
  \centering
  \includegraphics[width=\columnwidth,clip=yes,trim=0in 4.6in 4in 0in]{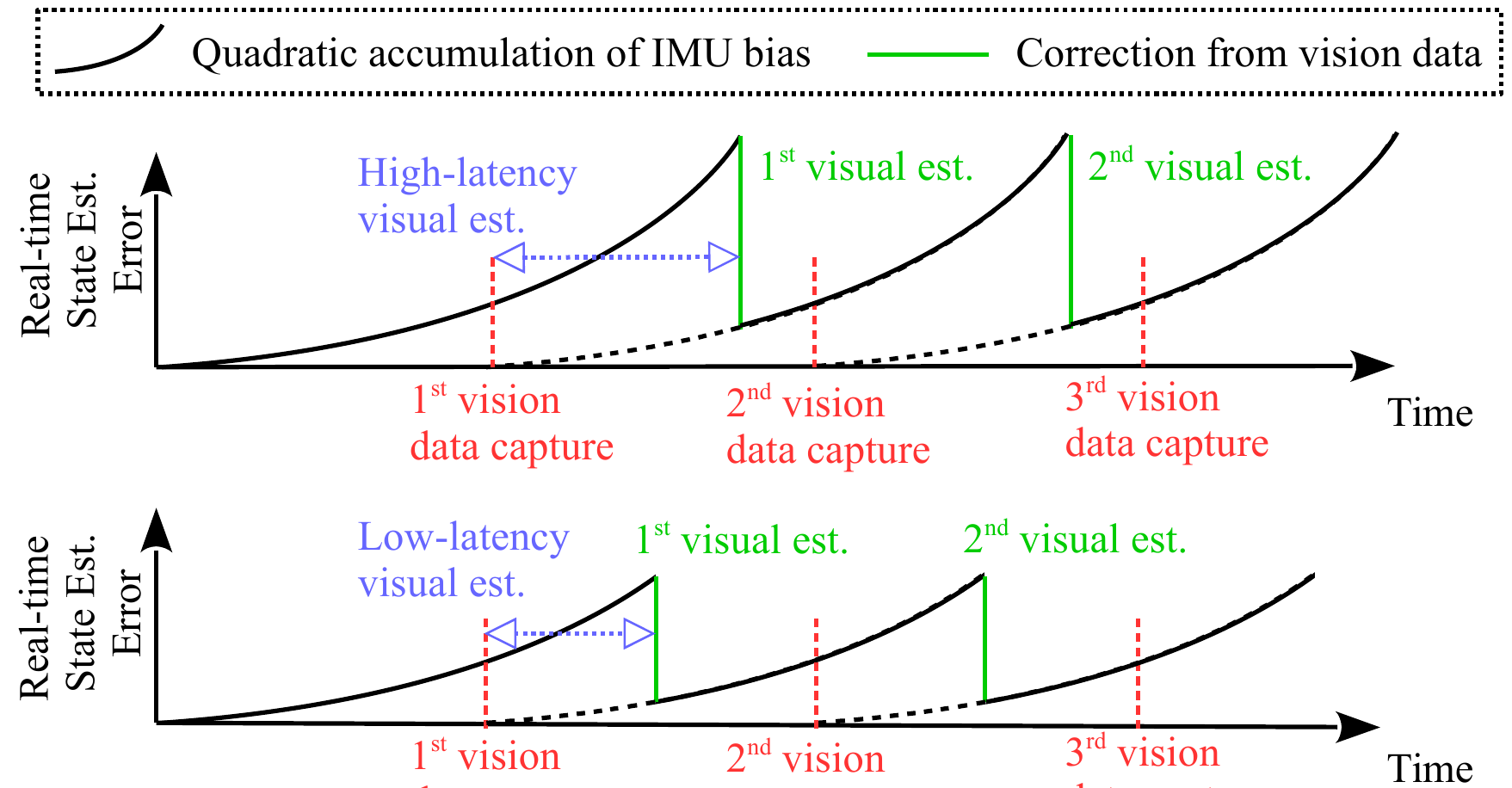}
  \caption{Impact of visual processing latency in VI-SLAM (best viewed in color).  
    Assuming 100\% correct visual estimation and purely-random IMU noise, 
    the only source of error in visual-inertial state estimation is 
    accumulated IMU bias (quadratic in time).  
  \textbf{Top: } trend of visual-inertial state estimation error when visual 
    estimation takes 75\% of the visual processing budget.
  \textbf{Bottom: } same error trend when visual 
    estimation takes 50\% of the budget.  
    Reduced latency yields a reduced state estimation error.  
  \label{fig:ImpactLatency}} 
  \vspace*{-1.5em}
\end{figure}

Though VI-SLAM drift is a critical factor influencing closed-loop 
navigation performance, 
the latency of visual estimation may also play an important role
when in closed-loop.
As illustrated in Fig.~\ref{fig:ImpactLatency}, latency-reduction on the
visual processing sub-system could improve the accuracy of fused
visual-inertial state estimate due to the quadratic (in time) nature of
accumulated IMU bias.  
Therefore,  this paper studies the impact of both drift and visual
estimation latency of VI-SLAM with closed-loop benchmarking simulation,
by implementing and testing several published VI-SLAM systems with
different run-time properties.  The closed-loop benchmarking outcomes
suggest that VI-SLAM systems must balance drift and latency.  
\section{Related Works}
%
This section first reviews existing works on visual-inertial state
estimation for closed-loop navigation.  The term \textbf{VI-SLAM} will
be used to indicate both visual-inertial odometry (VIO) and
visual-inertial SLAM.  After, it reviews evaluation methods for VI-SLAM
with a discussion of benchmarking for closed-loop trajectory tracking. 

\subsection{VI-SLAM in Closed-Loop Navigation}
There is a long history of using filters 
in visual-inertial state estimation for mobile robots 
(e.g., EKF \cite{howard2008real}; MSCKF \cite{mourikis2007multi}; 
\cite{cvivsic2018soft}).  
The combination of sparse optical flow (e.g., KLT \cite{GoodFeaturesToTrack}) 
and MSCKF has been recognized as an efficient VI-SLAM solution 
\cite{loianno2017estimation,sun2018robust,paschall2017fast}.  
A downside of most filter-based methods is the low mapping quality, 
which affects long-term navigation with location revisits.  

VI-SLAM running Bundle Adjustment (BA) retains an explicit map,
which promotes higher accuracy and long-term robustness of state
estimation.  To bound the cubic computational cost 
of BA, BA-based VI-SLAM typically works with a subset of 
historical information (keyframes and landmarks) 
sub-selected using a sliding window \cite{sibley2010sliding} 
or a covisibility graph \cite{strasdat2011double}.  
Representative BA-based VI-SLAM includes feature-based OKVIS 
\cite{leutenegger2015keyframe}, KLT-based VINS-Fusion \cite{qin2019a} 
and Kimera \cite{Rosinol19arxiv-Kimera}.   
Closed-loop navigation with OKVIS has been demonstrated on both ground 
\cite{blochliger2018topomap} and aerial robots \cite{burri2015real}.  
Full navigation has been demonstrated with VINS-Fusion on a micro-air
vehicle (MAV) \cite{lin2018autonomous}.  
Kimera \cite{Rosinol19arxiv-Kimera} estimates 3D mesh on-the-fly, which benefits navigation.

\begin{figure}[tb]
\vspace*{0.06in}
  \centering
  \includegraphics[width=0.75\columnwidth]{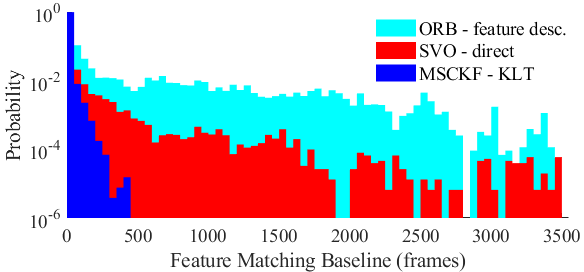}
  \vspace*{-0.4em}
  \caption{Distributions of feature matching baselines for 3 commonly used image 
  processing front-ends as computed for the EuRoC MAV benchmark 
  \cite{burri2016euroc}: 
  feature descriptors in ORB-SLAM (ORB) \cite{ORBSLAM}, 
  KLT in MSCKF \cite{sun2018robust}, and direct SVO \cite{SVO2017}.  
  For each feature/patch, the baseline is assessed by the {\em length of life}: 
  from the first-measured frame to the last-measured frame.  
  The feature-based front-end (ORB) extracts more long-baseline feature
  matchings than the KLT and direct methods.
  \label{fig:Baselines}} 
  \vspace*{-1.7em}
\end{figure}

Recently direct VI-SLAM systems have been derived; they do not require 
explicit feature extraction and matching. 
Direct systems jointly solve data association and state estimation 
by optimizing an objective functional using raw image readings.  
Direct VI-SLAM systems such as SVO \cite{SVO2017} and ROVIO 
\cite{bloesch2017iterated} have been integrated into closed-loop navigation 
systems \cite{papachristos2017autonomous,oleynikova2018complete}. 
While both KLT and direct VI-SLAM are computationally cheaper than
feature-based VI-SLAM, they are more sensitive to navigation-based
conditions: e.g. they require accurate pose prediction (from inertial) and 
minimal light condition changes.  Furthermore, both KLT and direct methods  
are mostly characterized by short-baseline feature matches.  Feature 
descriptor matching, on the other hand, can find reliable long-baseline 
feature matches for improved state estimation (see Fig.~\ref{fig:Baselines}). 

\subsection{Evaluation of Closed-Loop Trajectory Tracking}
Open-loop evaluation of different VI-SLAM methods has been extensively 
conducted in the literature, e.g. on multiple datasets 
\cite{bodin2018slambench2,zhao2020gfm}, on multiple computation devices 
\cite{BuEtAl_SLAMBench3[2019],delmerico2018benchmark,saeedi2018navigating}, 
and for multiple synthetic environments 
\cite{antonini2018blackbird,li2018interiornet}.
Closed-loop evaluation of different VI-SLAM methods in navigation tasks, 
however, is investigated less.  On-board evaluation tends to be reported
for individual implementations \cite{WeEtAl_VIOSwarm[2018],cvivsic2018soft}.  
One challenge of closed-loop evaluation is that closed-loop navigation
is not just a software problem; the performance of the full system is
affected by sensor choice, computational resources, system dynamics, and
target environment.  All these factors need to be experimentally
controlled to comprehensively evaluate the performance of closed-loop
navigation using VI-SLAM. 

One way to conduct comprehensive and repeatable closed-loop evaluation
is via simulation.  Several existing simulators are commonly used in the
robotics community.  
Gazebo \cite{koenig2004design} is one of the most popular simulators, 
with MAV-specific extensions such as RotorS \cite{furrer2016rotors}.
AirSim \cite{shah2018airsim} is another choice, 
with photorealistic renderings of visual data via Unreal Engine.  
A more recent development incorporates hardware in the 
loop \cite{sayre2018visual}.  
The approach captures the trajectory of the actual robot on the fly,
while rendering virtual visual data on a remote workstation to
collect actual data under real physics and virtual data from an
easy-to-extend renderer.  
However, ground truth acquisition relies on a MoCap device, 
which is hard to scale beyond room-sized environment.
To properly benchmarking VI-SLAM in closed-loop navigation, the 
benchmarking framework needs to be re-configurable to cover a variety of 
factors, such as sensor configurations, computational \& robot platforms, 
and target environments.  
Furthermore, ground truth coverage is required over the entire course 
of navigation.  This work aims to fill a existing gap by presenting 
an open-source, closed-loop benchmarking framework that supports the
above requirements, and serves to provide performance insights on
representative VI-SLAM systems based on the closed-loop evaluation
results.  


%
\section{Closed-Loop System Overview}
\begin{figure}[!tb]
\vspace*{0.06in}
  \centering
  \includegraphics[width=0.96\columnwidth,clip=false,trim=0.2in 4.85in 1.7in 0in]{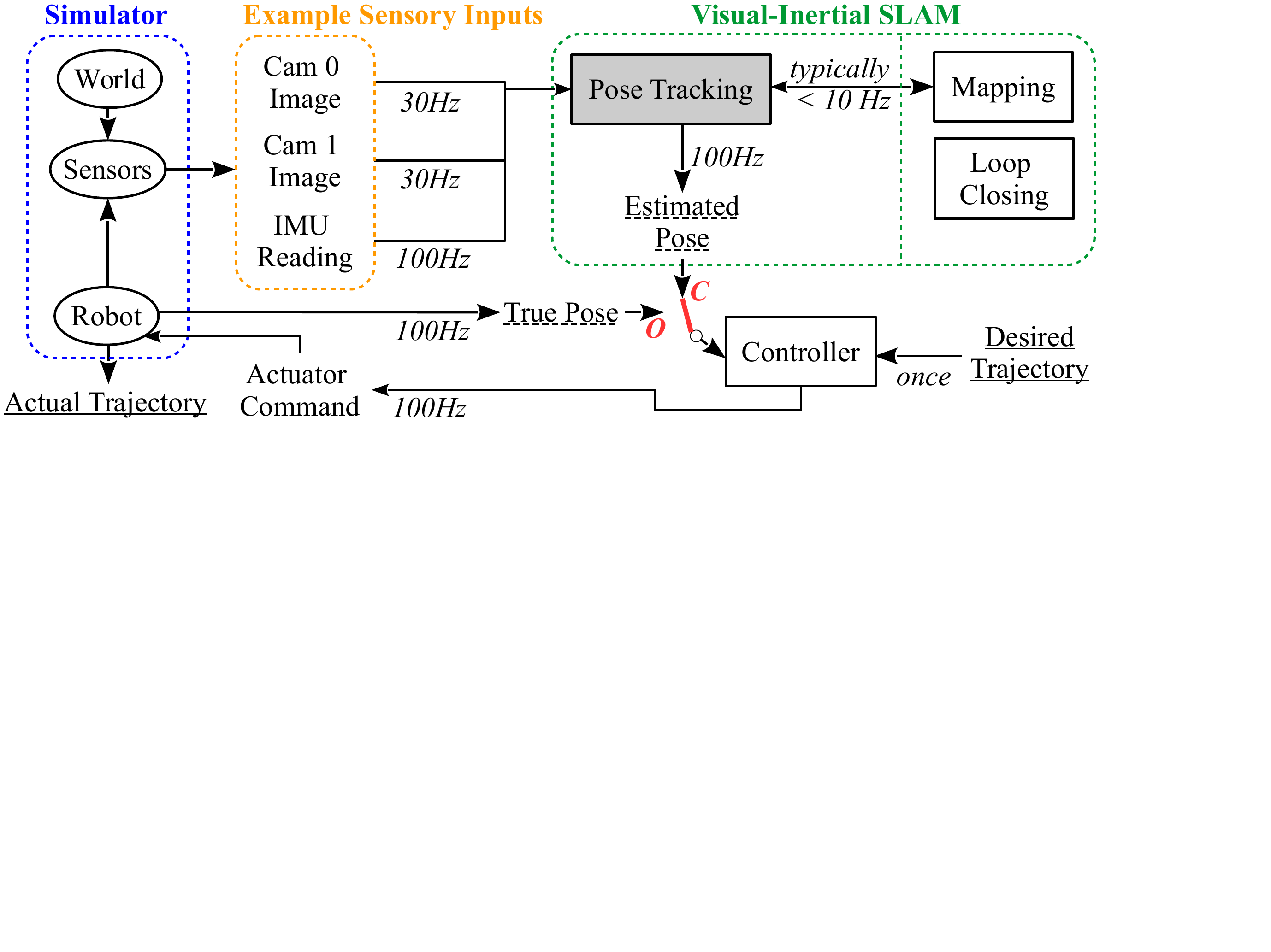}
  \caption{Overview of the closed-loop trajectory tracking system.  
  The Gazebo simulator sends out sensory data to VI-SLAM.  
  The pose-tracking module of VI-SLAM processes the data, and
  outputs high-rate pose estimation.  
  In closed-loop benchmarking (red switch at ``C''), the pose estimation
  is taken by the controller to generate high-rate actuator command,
  which is sent back to the simulated robot in Gazebo.  Performance
  is quantified by comparing desired and actual trajectories (solid underlined
  text).  
  In open-loop benchmarking (red switch at ``O''), the controller
  generates commands based on the true pose, available from the
  simulator.  Performance is quantified by comparing true and estimated poses
  (dashed underlined text).\label{fig:Overview}} 
  \vspace*{-1.5em}
\end{figure}

The closed-loop trajectory tracking system consists of two major
subsystems, illustrated in Fig.~\ref{fig:Overview} and described as: 
1) a VI-SLAM system taking vision \& inertial data to generate 
  high-rate state estimates and low-rate map updates; and 
2) a controller using high-rate output from the pose tracking module 
of VI-SLAM to generate actuator commands. 
%
Though this paper covers only stereo-inertial sensory inputs, 
the system supports other common visual sensors such as monocular 
and RGB-D cameras.  

While both mapping and loop closing are essential for accurate and robust
state estimation, these two modules require high computation, and typically 
operate at a much lower rate than pose tracking (usually by an order of 
magnitude).  
Therefore the high-rate pose estimation required in feedback control is 
collected by the pose tracking module.  
This study explores the efficiency and accuracy of the pose tracking 
module when used for feedback.  
A variety of VI-SLAM systems are integrated into the 
closed-loop system, covering representative design options such as 
loosely/tightly-coupled visual-inertial fusion,  
direct/feature-based data association, and 
filter/BA-based back-end.

The focus of this study is the trajectory tracking performance 
of VI-SLAM systems when used in the closed-loop.  
The tracking performance can be reflected by computing the 
pose error between the desired and actual trajectories,
accumulated over the entire course of navigation.  
Here we report the root mean square of the translation error between the
desired and actual trajectories, dubbed tracking RMSE, as the
performance metric.  
Tracking RMSE matches the formulation of ATE \cite{sturm12iros_ws},
commonly used in open-loop evaluation, but works with actual robot
trajectory.  It directly measures the end performance of the trajectory
tracking system, thereby capturing the joint effect of pose tracking
drift and latency.  Additional metrics that capture orientation error 
are reported online (see \cite{ZhEtAl_SLAM_CLBenchGit}).


The mobile robot used in the simulation is the differential drive
TurtleBot2.  Mounted to the robot are a 30fps stereo camera with an 11cm 
baseline, and an IMU placed at its base.  Data streams from both the stereo
camera and IMU are input to the VI-SLAM system, which outputs $SE(3)$ state 
estimates.  The trajectory tracking controller uses the $SE(2)$ subspace
of the $SE(3)$ estimate to track the target trajectory.  
The next subsections describe the implemented VI-SLAM systems, the
trajectory tracking controller, and the simulation setup in Gazebo/ROS.

\subsection{Visual-Inertial SLAM Systems}
Several publicly available stereo(-inertial) SLAM implementations were
selected for integration into the closed-loop benchmarking system.  
The five implementations are:
\setlength{\IEEEelabelindent}{0em}%
\begin{enumerate}
\item \textit{MSC}: MSCKF-VIO \cite{sun2018robust} + MSF \cite{lynen2013robust}.
  MSCKF-VIO is a tightly-coupled VIO system, with KLT-based front-end and MSCKF back-end.  
  EKF-based sensor fusion, MSF \cite{lynen2013robust}, densifies the 
  low-rate estimation output from MSCKF-VIO, before sending it to controller.
  No loop closing is included.  
  %
\item \textit{VINS}: VINS-Fusion \cite{qin2019a}
  is a tightly-coupled VI-SLAM with KLT-based front-end and BA-based
  back-end.  \textit{VINS} has a large latency due to the BA.  
  It does provide a low-latency, high-rate IMU propagation signal, 
  which is sent to controller.  \textit{VINS} comes with loop closing, which
  is preferred in long-term revisit scenarios.  
  A circular motion is executed to initialize \textit{VINS} prior to starting
  the SLAM estimation process.
  %
\item \textit{SVO}: SVO \cite{SVO2017} + MSF.  
  An efficient, loosely-coupled VIO system that consists of direct SVO 
  and MSF fusion.  No loop closing module.  \textit{SVO} has the lowest pose tracking 
  latency of the methods listed.
  %
\item \textit{ORB}: ORB-SLAM \cite{ORBSLAM} + MSF.  
  ORB-SLAM has a feature-based front-end and BA-based back-end.  
  Due to the computational-costly feature extraction and matching, ORB-SLAM 
  has a large visual estimation latency.  
  Similar with \textit{MSC} and \textit{SVO}, MSF is integrated into {ORB} to 
  generate a high-rate estimation signal.
%
\item \textit{GF}: Lazy-GF-ORB-SLAM \cite{zhao2020gfm} + MSF.
  A loosely-coupled modified version of \textit{ORB} with two
  efficiency modifications: good feature and lazy stereo.  
  Good feature matching performs targeted map-to-frame
  matching under an upper bounded matching budget. The lazy stereo
  modification partitions the stereo ORB-SLAM computations into those
  necessary for immediate pose estimation versus those that assist future
  pose estimation computations. The former is prioritized to run first, 
  therefore enabling more rapidly output of pose estimation. These 
  two modifications lower the latency without significant impact 
  on the accuracy of pose estimation.
\end{enumerate}
If no initialization approach is described, then
the default is to keep the robot static for 10 seconds before starting a
closed-loop/open-loop run.

\subsection{Feedback Control}
\newcommand{\position}{\ensuremath{d}}
\newcommand{\roboPose}{\ensuremath{g}}
\newcommand{\velocity}{{v}}
\newcommand{\velocities}{{V}}
\newcommand{\accel}{{u}}

The desired trajectory $\position^*(t) \in \mathbb{R}^2$ is constructed from a
series of specified waypoints using splines.  An exponentially
stabilizing trajectory tracking controller for Hilare-style robots
\cite{near-identity} generates a kinematically feasible trajectory for the
robot to follow.  In the following discussion, constraints on accelerations
and velocities are omitted for clarity, though they exist within the actual
implementation.

The robot pose as a function of time $\roboPose(t) \in SE(2)$ obeys
follows the control equations: 
\begin{equation}	\label{eqn:nearidentity-robotstate}
  \dot \roboPose 
    = \roboPose \cdot \begin{bmatrix} \nu \\ 0 \\ \omega \end{bmatrix} 
  \quad
  \text{and}
  \quad
  \begin{matrix}
    & \dot \nu    = \accel^1 \\
    & \dot \omega = \accel^2 
  \end{matrix}
\end{equation}
where $\nu$ is the forward velocity and $\omega$ is the angular
velocity, both in the body frame.
The signal $u = (u^1, u^2)^T$ coordinates are the forward and angular
acceleration (in body frame).

The controller used relies on the differential flatness of the robot
motion to achieve exponential stabilization of a virtual point in front
of the robot (by a distance $\lambda$) \cite{near-identity}.
Define the $\lambda$-adjusted rotation matrix and
angular velocity matrix,
\begin{equation}	\label{eqn:nearidentity-adjusted}
  R_{\lambda} = R \cdot \diag(1, \lambda) 
  \quad \text{and} \quad
  \hat{\omega}(\lambda,\dot{\lambda}) = 
    \begin{bmatrix}
      0 & -\lambda \omega \\
      \frac{1}{\lambda}\omega & \frac{\dot{\lambda}}{\lambda}
    \end{bmatrix},
\end{equation}
where $R$ is the rotation matrix given by the orientation in $\roboPose$.
For $e_1$ the unit body $\hat x$-vector in the world frame, the trajectory
tracking control law is
\begin{multline}	\label{eqn:nearidentity-controlsignal}
  \accel = c_{p}R^{-1}_{\lambda} \of{\position^* - \position - \lambda*Re_{1}}
             + c_d \of{ R^{-1}_{\lambda} \dot{\position}^* - \velocities} \\
           - c_{d}\dot{\lambda}e_{1} 
             - \hat{\omega}(\lambda,\dot{\lambda})\velocities 
             - (\hat{\omega}(\lambda,\dot{\lambda}) 
                            - c_{\lambda}I) \dot{\lambda}e_{1},
\end{multline}
where $c_{p}, c_d, c_{\lambda}$ are feedback gains 
and $\velocities=\big[\nu\, | \,\omega \big]^{T}$. 
The additional offset dynamics are
\begin{equation}	\label{eqn:nearidentity-offsetdynamics}
  \dot{\lambda} = -c_{\lambda}(\lambda - \epsilon), \ \text{where}\ 
    \lambda(0) > \epsilon > 0,\  c_{\lambda} > 0.
\end{equation}
The dynamical system represented by
Eqs~\ref{eqn:nearidentity-robotstate}-\ref{eqn:nearidentity-offsetdynamics}
yields a reference trajectory of robot poses $g^*(t)$ and body velocity
components $\velocities^*(t)$ for tracking the desired trajectory $\position^*(t)$.
The offset variable $\lambda^*(t)$ can be ignored.
%

The real time trajectory controller drives the robot to track the
reference trajectory based on feedback of the robot's estimated state 
(an $SE(2)$ substate of the $SE(3)$ state estimate).
These control commands are:
\begin{equation}
\begin{split}
\nu_{cmd} & = k_{x}*\widetilde{x} + \nu^{*} \\
w_{cmd} &= k_{\theta}*\widetilde{\theta} + k_{y}*\widetilde{y} + \omega^{*} \\
\end{split}
\end{equation}
where $[\widetilde{x},\widetilde{y},\widetilde{\theta}]^T \simeq
\widetilde{g} = \inverse{\roboPose} \roboPose^*$ 
is the relative pose error between the current state $g$ and the 
desired state $g^{*}$ in body frame.
In the absence of error, the control signal is $\velocities^{*}(t)$.

\subsection{Simulation Setup}

\begin{figure}[tb]
\vspace*{0.06in}
  \centering
  \includegraphics[width=0.8\columnwidth]{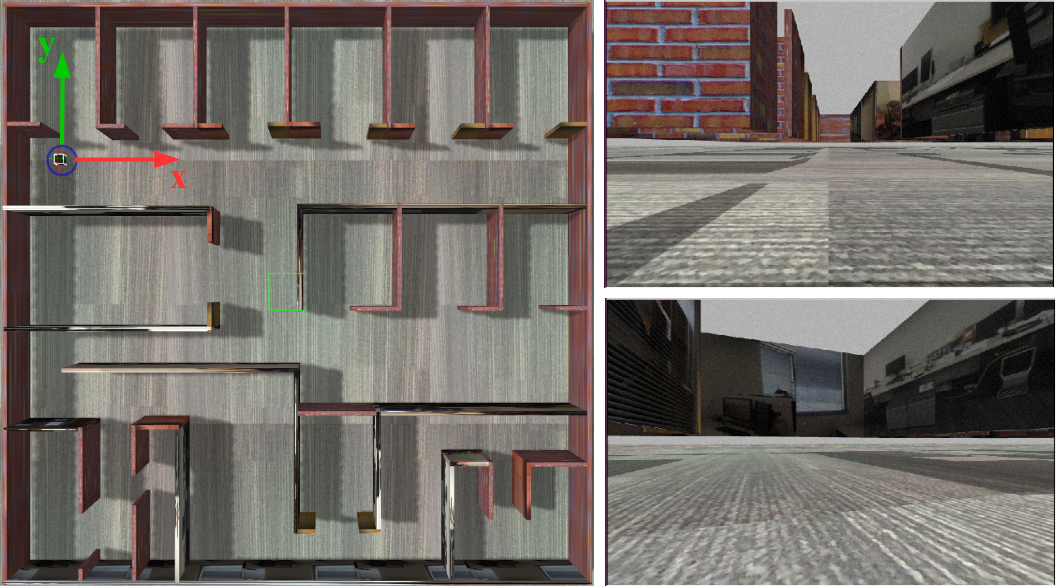} 
  \vspace*{-0.10in}
  \caption{The virtual office world.
  \textbf{Left: } Top-down view.  The robot starts at the top-left
    corner, facing the long corridor.
  \textbf{Right: } Example images captured by on-board stereo camera 
    (left camera). 
  \label{fig:SimuWorld}} 
  \vspace*{-0.2in}
\end{figure}
This section describes the Gazebo-simulated environment for testing
closed-loop trajectory tracking with VI-SLAM systems.  
The scene created for robot navigation is a virtual office world 
(Fig.~\ref{fig:SimuWorld}).  
The world is based on the floor-plan of an actual office, with texture-mapped
surfaces.  The walls are placed 1m above the ground plane since collision
checking and path planning is outside the scope of this paper.  
Introducing collision avoidance would add another coupling factor to the
closed-loop system, which would introduce unneeded difficulty in
identifying the source of tracking error (i.e., was it to avoid a
collision or due to poor estimation?).

\begin{figure}[t]
\vspace*{0.06in}
  \centering
  \includegraphics[width=\columnwidth,clip=false,trim=0in 0.30in 0in 0in]{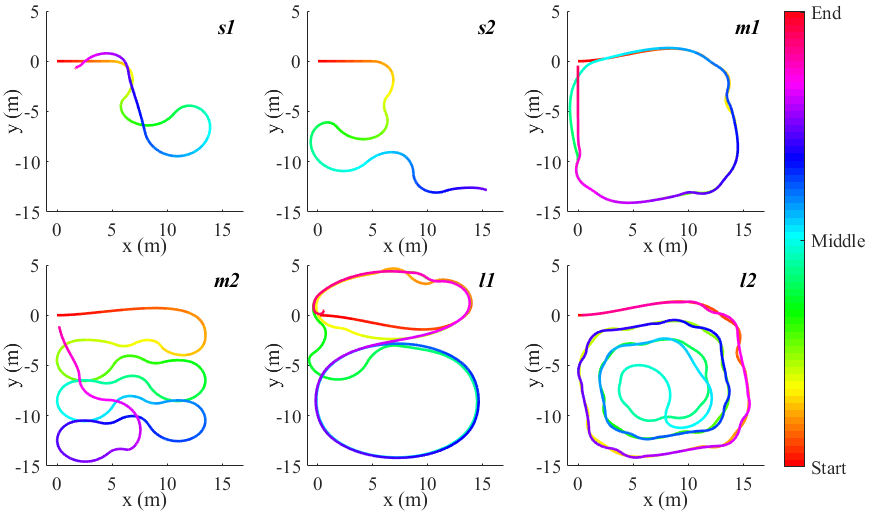} 
  \caption{All 6 desired trajectories used in closed-loop navigation experiments. 
  Each desired trajectory is color-coded to show the direction of travel.
  \label{fig:Paths}} 
  \vspace*{-2em}
\end{figure}

Six test trajectories were created for the closed-loop navigation experiments, 
each with different characteristics (Fig.~\ref{fig:Paths}).  
The first two are relatively short ($\sim$50m), with few to no revisits.
The 3rd and 4th trajectories are both of medium length ($\sim$120m). 
The 3rd has many revisits as it retraces the trajectory once, whereas
the 4th crosses earlier trajectory segments facing the opposite direction or 
transverse to them.
The last 2 trajectories are long ($\sim$240m). 
The 5th retraces trajectory segments, while the 6th does so facing in
the opposite direction.
All trajectories have the same start point for the robot, the origin of
the world.  Three desired linear velocities are tested: 0.5m/s, 1.0m/s,
and 1.5m/s.  Based on these velocities, the navigation course in
simulation lasted from 30 seconds up to 480 seconds.

%
\section{Experimental Results} 
This section describes the outcomes of two main experiments. The first
involves open-loop evaluation of the stereo VI-SLAM methods, where  
the controller takes true poses instead of VI-SLAM estimations 
(red ``O'' in Fig.~\ref{fig:Overview}). 
The open-loop evaluation serves two purposes: 
1) it demonstrates that the relative ranking of the methods in the
simulated world is roughly preserved when compared to video-recorded
open-loop benchmarks such as EuRoC \cite{burri2016euroc}, and 
2) it characterizes the functional domain of the simulation environment
relative to the benchmarks.  
The second experiment performs closed-loop trajectory tracking tests,
where the controller is fed with VI-SLAM estimation (red ``C'' in
Fig.~\ref{fig:Overview}).  The objective of closed-loop benchmarking is
to identify the VI-SLAM properties (i.e., drift, latency) that have
impact on trajectory tracking performance.  

All 5 VI-SLAM systems parameter configurations were found via 
parameter sweep.  For each test configuration (desired trajectory,
desired linear velocity, VI-SLAM method, and IMU), 
the benchmarking run is repeated five times, so that 
random factors such as multi-threading and 
random sensory noise are properly reflected.  
Two commonly-used IMUs are simulated:
a high-end ADIS16448 and a low-end MPU6000.  
Open-loop benchmarking was performed on an i7-4770 
(single thread Passmark score: 2228). 
Closed-loop benchmarking was performed on a dual Intel Xeon E5-2680 workstation 
(single thread Passmark score: 1661).  
For reference, most published closed-loop navigation systems 
\cite{scaramuzza2014vision,burri2015real,paschall2017fast,cvivsic2018soft,sun2018robust,lin2018autonomous,oleynikova2018complete} 
employ an Intel NUC whose CPUs score between 1900-2300 (single thread). 
The full stack, including the simulator, integrated VI-SLAM systems and 
trajectory tracking controller, are released \cite{ZhEtAl_SLAM_CLBenchGit}. 

\subsection{Open-Loop Outcomes and Analysis} 
\begin{table}[!tb]
\vspace*{0.06in}
  \footnotesize
  \centering
  \caption{Characterization of Benchmarks \label{charseq}}
    \vspace*{-0.5em}
  \begin{tabular}{|c||cccc|}
    \hline
    \textbf{Seq.} & Duration & Motion & Revisit & Feats.\\ \hline
    \textit{MH 03 med}   & medium & medium & high & 235 \\
    \textit{MH 04 diff}  & short  & high   & high & 235 \\
    \textit{VR1 01 easy} & medium & medium & high & 227 \\
    \textit{VR2 02 med}  & short  & medium & high & 246 \\
    \textit{MH 05 diff}  & short  & high   & high & 240 \\ \hline
    \textit{s1}          & short  & low-high & low & 131 \\
    \textit{s2}          & short  & low-high & low & 126 \\
    \textit{m1}          & short  & low-high & high & 122 \\
    \textit{m2}          & medium & low-high & low & 115 \\
    \textit{l1}          & medium & low-high & high & 121 \\
    \textit{l2}          & medium & low-high & med  & 114 \\ \hline
  \end{tabular}
    \vspace{-.5cm}
\end{table}

Given that simulation and recorded open-loop benchmark data may not 
align, this section conducts a comparison for verification of the domain
of applicability for the simulation. 
The comparison shows that simulated scenes have some overlap with existing
benchmarks though they do not span the entire domain. Based on the
similarity, closed-loop implementations should have predictive power 
when the closed-loop system is deployed in equivalent real-world
settings.

\begin{table*}[!tb]
\vspace*{0.06in}
  \centering
  \begin{minipage}{0.475\textwidth}
  \centering
  \caption{Open-loop Outcomes on EuRoC (ATE in m; Latency in ms)
    \label{tab:BMVid_OpenLoop}}
    \vspace*{-0.5em}
  \begin{tabular}{|c||ccccc|}
    \hline
    \textbf{Seq.}    & SVO  & MSC & GF & ORB & VINS \\
    \hline
    \textit{MH 03 med}   & 0.31 & 0.24 & 0.07 & {0.05} & 1.50 \\
    \textit{MH 04 diff}  & 2.78 & 0.28 & {0.10} & 0.15 & 2.24 \\
    \textit{VR1 01 easy} & 0.05 & 0.12 & 0.10 & {0.04} & 0.35 \\
    \textit{VR2 02 med}  & 0.19 & 0.22 & {0.04} & 0.09 & 0.35 \\
    \textit{MH 05 diff}  & 0.47 & 0.28 & {0.04} & 0.20 & 2.32 \\ 
    & & & & & \\ \hline
    \textbf{Avg. ATE}     & 0.76 & 0.23 & {0.07} & 0.11 & 1.35 \\ \hline \hline
    \textbf{Avg. Latency}  & {16.4} & 28.3 & 20.7 & 38.5 & 93.5 \\ \hline
  \end{tabular}
  \end{minipage}
  \begin{minipage}{0.475\textwidth}
  \centering
  \caption{Open-loop Simulation Outcomes (ATE in m; Latency in ms)
    \label{tab:BMSim_OpenLoop}}
    \vspace*{-0.5em}
  \begin{tabular}{|c||ccccc|}
    \hline
    \textbf{Seq.}    & SVO  & MSC & GF & ORB & VINS \\
    \hline
    \textit{s1} & 0.15 & 0.21 & {0.12} & 0.14 & 0.14 \\
    \textit{s2} & 0.12 & 0.14 & {0.11} & 0.38 & 0.13 \\
    \textit{m1} & 0.33 & 0.32 & 0.19 & {0.16} &  --  \\
    \textit{m2} & 0.45 & 0.29 & {0.19} & 0.20 & 0.33 \\
    \textit{l1} & 0.42 & 0.57 & 0.25 & {0.09} & 0.57 \\
    \textit{l2} & - & 0.51 & {0.29} & 0.38 & 0.47 \\ \hline 
    \textbf{Avg. ATE}     & 0.29 & 0.34 & {0.19} & 0.23 & 0.32 \\ \hline \hline
    \textbf{Avg. Latency} & 9.3 & 14.2 & 26.2 & 47.5 & 62.0 \\ \hline
  \end{tabular}
  \end{minipage}
\end{table*}

A subset of the EuRoC sequences and the simulated open-loop sequences 
are characterized in Table \ref{charseq}, using the benchmark properties 
evaluated in \cite{ye2019characterizing}. 
The duration profile is defined with medium describing an interval
of $[2,10]$ minutes. 
The motion profile is categorized from \textit{low} (0.5
m/s) to \textit{medium} (1.0 m/s) to \textit{high} (1.5 m/s).
The revisit frequency is a function of the trajectory followed and how
often there is co-visibility of features across trajectory segments that are
temporally distant. 
One additional statistic
captured is the average number of features tracked per frame using \textit{ORB} 
(last column). 
Simulation sequences exhibit less texture than EuRoC ones.  
There is sufficient overlap in the characteristics of the two
benchmarks, with the simulation reflecting slightly more diverse
scenarios.  Qualitatively, the simulation sequences are comparable or
harder benchmarking cases relative to the EuRoC sequences.

To compare further, we ran open-loop benchmarking against ground truth to
get a sense for the pose estimation properties of the VI-SLAM
algorithms and whether the two benchmark sets agree in terms of relative
ordering.  The results averaged from 5-run repeats of the \textit{medium}
motion profile are summarized in Tables \ref{tab:BMVid_OpenLoop} and
\ref{tab:BMSim_OpenLoop}, where the track loss cases are omitted
(dashes).  
According to the tables, the ATE between VI-SLAM
estimation and ground truth is usually lower for EuRoC sequences. Both
SVO and VINS exhibit outliers in EuRoC relative to the prevailing values
across all methods, with SVO having one and VINS having three. They
occur for the MH sequences, suggesting that these might be more
problematic in general for VINS and SVO. However, SVO and VINS also have
one track failure for the simulated cases.  Overall, the outcomes align
with the previous claim that the simulation sequences are
comparable to or harder than EuRoC.
If track failure is added as a penalty to the simulation performance
outcomes, then the rank ordering of the algorithms agree between EuRoC
and simulation. 
{\em GF} typically 
has the lowest ATE, while {\em VINS} has the highest ATE. 
Furthermore, the relative orders of latencies for different VI-SLAM 
agree: {\em SVO} is lowest, and {\em VINS} is highest 
(MSC is unique in that is has mismatch).
The comparisons support using simulation to benchmark VI-SLAM, with
validity for specific deployment conditions.

\subsection{Closed-Loop Outcomes and Analysis} 


Trajectory tracking performance is quantified in 
Tables~\ref{tab:Nav_RMSE_High} and \ref{tab:Nav_RMSE_Low}.  
The tracking RMSE between the desired and actual trajectories 
reflects an average of the 5-run repeats.  
Cases with average RMSE over 10m are considered navigation
failures and omitted (dashes). 
%
The average latency of visual estimation in each VI-SLAM is reported in
the bottom row of each table (algorithms sorted to be in ascending order).

\begin{table*}[tb!]
  \centering
  \caption{Closed-loop Outcomes with High-end IMU ADIS16448 (Tracking RMSE in m; Latency in ms)}
  \label{tab:Nav_RMSE_High}
  \vspace*{-0.5em}
  \addtolength{\tabcolsep}{-2pt}
  \begin{tabular}{|c||ccccc||ccccc||ccccc|}
	\hline 
	\textbf{ } & 
	\multicolumn{5}{c|}{\bfseries \small 0.5m/s} & 
	\multicolumn{5}{c|}{\bfseries \small 1.0m/s} & 
	\multicolumn{5}{c|}{\bfseries \small 1.5m/s} \\
	\textbf{\small Seq.} 
	& SVO & MSC & GF  & ORB & VINS
	& SVO & MSC & GF  & ORB & VINS
	& SVO & MSC & GF  & ORB & VINS \\
	\hline
	\textit{s1} 
    & 0.23 & 0.65 & {0.11} & 0.24 &  --  
    & 0.56 & 0.26 & {0.12} & 0.28 & 1.36 
    & 0.49 & 0.22 & {0.14} & 0.23 & 0.37 \\
 	\textit{s2} 
 	& 0.18 & 0.46 & {0.09} & 0.43 &  --
    & 1.13 & 0.38 & {0.08} &  --  &  --
    & 1.21 & 0.33 & {0.09} & 3.26 &  -- \\
	\textit{m1} 
    & 0.92 & 1.54 & {0.12} & 0.31 &  -- 
    &  --  & 1.01 & {0.10} & 0.23 &  -- 
    & 1.26 & 0.81 & {0.11} & 2.10 &  -- \\
	\textit{m2}	
    & 0.36 & 2.23 & {0.14} &  --  &  -- 
    & 0.86 & 1.53 & {0.12} &  --  &  --
    & 1.87 & 0.68 & {0.14} &  --  &  -- \\
	\textit{l1}	
    & {2.12} & 2.73 &  --  &  --  &  --
    & 1.79 & 6.67 & {0.15} &  --  &  --
    & 1.22 & 2.13 & {0.22} &  --  &  -- \\
	\textit{l2}	
    & 0.87 & 2.62 & 0.36 & {0.24} &  -- 
    & 1.27 & 3.25 & 0.35 & {0.31} &  -- 
    & 2.78 & 2.66 & {0.35} & 0.37 &  -- \\
	\hline	
	\textbf{\small Avg. RMS} 
    & 0.78 & 1.70 & {0.16} & 0.30 &  -- 
    & 1.12 & 2.18 & {0.15} & 0.27 & 1.36 
    & 1.47 & 1.14 & {0.18} & 1.49 & 0.37 \\
	\hline	
	\hline
	\textbf{\small Avg. Latency} 
	& 8.9 & 17.7 & 32.8 & 52.4 & 55.0 
	& 8.9 & 16.9 & 32.4 & 51.7 & 73.9 
	& 8.9 & 16.7 & 32.0 & 50.6 & 64.1   \\ 
		\hline	
	\end{tabular}
	\caption{Closed-loop Outcomes with Low-end IMU MPU6000 (Tracking RMSE in m; Latency in ms)}
	\label{tab:Nav_RMSE_Low}
    \vspace*{-0.5em}
	\begin{tabular}{|c||ccccc||ccccc||ccccc|}
		\hline 
		\textbf{ } & 
		\multicolumn{5}{c|}{\bfseries \small 0.5m/s} & 
		\multicolumn{5}{c|}{\bfseries \small 1.0m/s} & 
		\multicolumn{5}{c|}{\bfseries \small 1.5m/s} \\
		\textbf{\small Seq.} 
		& SVO & MSC & GF  & ORB & VINS
		& SVO & MSC & GF  & ORB & VINS
		& SVO & MSC & GF  & ORB & VINS \\
		\hline
		\textit{s1} 
        & 0.68 & 0.29 & {0.13} & 0.05 & 0.52
        & 1.21 & 0.35 & {0.16} & 0.25 & 0.89
        & 2.30 &  --  & {0.41} & 0.46 &  -- \\
 		\textit{s2} 
        & 0.62 & {0.40} & 5.00 & 0.96 & --
        & 0.96 & 0.35 & {0.10} &  --  & --
        & 2.68 &  --  & {0.21} &  --  & -- \\
		\textit{m1} 
        & 1.53 & 0.68 & 0.26 & {0.19} & 8.24 
        & 3.21 &  --  & {0.28} & 1.21 &  --  
        & 3.95 &  --  & {0.36} &  --  &  -- \\
		\textit{m2}	
        & 2.13 & 1.60 & {0.43} &  --  &  --
        & 2.33 &  --  & {0.53} &  --  &  --
        & 4.21 &  --  & {0.39} &  --  &  -- \\
		\textit{l1}	
        & {0.18} & 4.60 & 3.24 & 1.62 &  --
        & 2.19 &  --  & {0.37} &  --  &  --
        & 2.46 & {1.87} & 3.41 &  --  &  -- \\
		\textit{l2}	
        & {0.21} & 3.74 & 0.36 &  --  &  --
        & 2.46 & 5.67 & {0.35} &  --  &  --
        & 1.86 & 1.92 & {0.32} &  --  &  -- \\
		\hline	
		\textbf{\small Avg. RMS} 
        & 0.89 & 1.88 & 1.57 & {0.71} & 4.38 
        & 2.06 & 2.12 & {0.30} & 0.73 & 0.89
        & 2.91 & 1.90 & 0.85 & {0.46} &  --    \\
		\hline	
		\hline
		\textbf{\small Avg. Latency} 
		& 10.1 & 17.4 & 28.9 & 53.2 & 62.3  
		& 10.0 & 16.9 & 31.5 & 53.0 & 65.4  
		& 10.0 & 16.5 & 31.8 & 52.8 & 58.8   \\ 
		\hline	
	\end{tabular} 
  \vspace*{-1.5em}
\end{table*}

\begin{figure*}[t]
  \vspace*{0.5em}
  {
  \begin{minipage}{\columnwidth}
    \centering
  \includegraphics[width=\columnwidth,clip=false,trim=0in 0.3in 0in 0in]{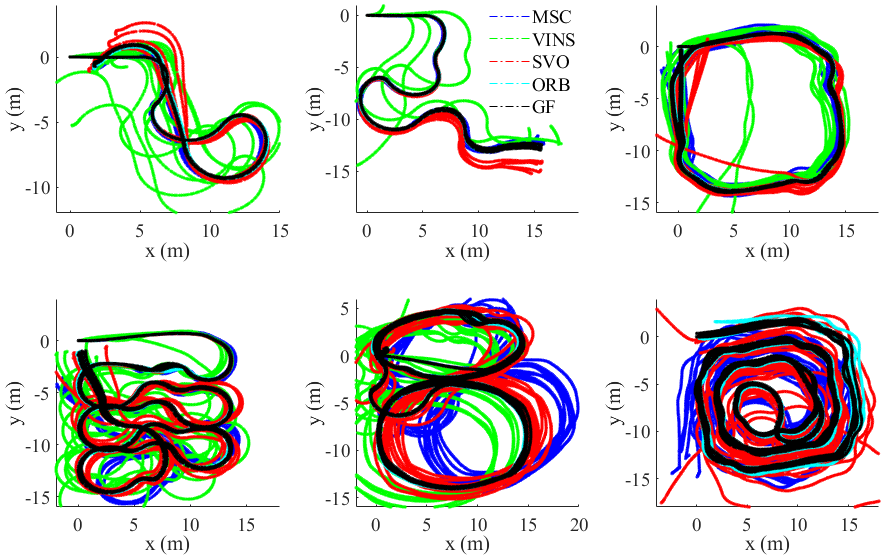}
  \caption{Actual trajectories the robot traveled for each desired trajectory, 
  color-coded by method.  Desired velocity is 1.0m/s and IMU is high-end.
  \label{fig:Nav_Track_High}} 
  \end{minipage}
  \hfill
  \begin{minipage}{\columnwidth}
    \centering
    \includegraphics[width=\columnwidth,clip=false,trim=0in 0.3in 0in 0in]{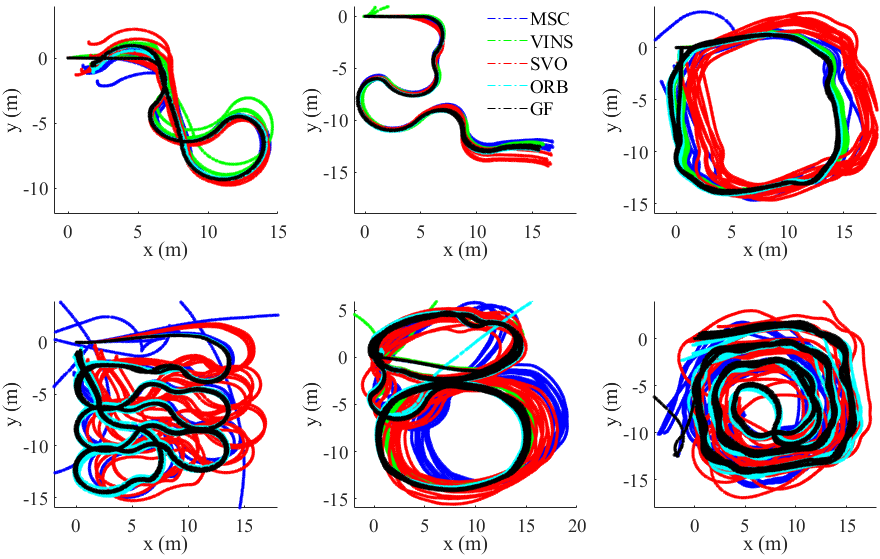}
    \caption{Actual trajectories the robot traveled for each desired trajectory, 
  color-coded by method.  Desired velocity is 1.0m/s and IMU is low-end.
    \label{fig:Nav_Track_Low}} 
  \end{minipage}}
  \vspace*{-1.0em}
\end{figure*}

%
According to Tables~\ref{tab:Nav_RMSE_High} and \ref{tab:Nav_RMSE_Low}, both \textit{VINS} and \textit{ORB} fail under
multiple configurations.  
Compared to {\em ORB}, the success rate and RMS of {\em GF} are significantly 
improved.  The reduction of visual estimation latency contributes to the 
improvement, since the open-loop outcomes of \textit{ORB} and \textit{GF} 
are similar in terms of drift, but are quite different in terms of latency.
The outcomes suggest that meeting standard frame-rate latency levels
($\sim\!\!30$ms) is best, and quite possibly essential for good
closed-loop trajectory tracking performance. 
%
%
The filter-based \textit{MSC} is significantly affected by the IMU data
quality, as it fails to navigate for multiple low-end IMU cases and higher
velocity. The outcomes suggest an over-reliance on the IMU for pose
estimation, which is supported by Fig.~\ref{fig:Baselines} where {\em MSC}
has poor long-term data association for detected features. 
Being able to
re-associate to lost tracks improves performance by linking against a known
static point in the world and improving absolute position estimates.
Otherwise, systems such as {\em MSC} rely on integrated estimates which have poor
observability properties.

%
%
%

%
The last two approaches to review are $SVO$ and $GF$, which both successfully
track the camera pose for all but one sequence, each. These are the two
strongest performing methods. Interestingly they have different run-time
properties. For both the open-loop and closed-loop evaluations, \textit{SVO} has the
lowest latency but the highest drift, while \textit{GF} is the opposite. Their
relative performance remained the same from open-loop to closed-loop,
modulo a small fraction of sequences. It appears that low latencies are
tolerant to higher drift, whereas lower drift permits higher latency.
Overall, however, it appears that once the latency is low enough, it is
better to target accuracy enhancements over latency enhancements for
closed-loop trajectory tracking 
(for ground vehicles in mostly static, feature sufficient, environments).
Comparing \textit{SVO} and \textit{GF} across the two IMU types indicates that high-end IMUs
provide the best error scaling to ground speed, with \textit{GF} being more
consistent as the speed increased.


These quantitative outcomes can be seen qualitatively in 
Fig.~\ref{fig:Nav_Track_High} and \ref{fig:Nav_Track_Low}, which trace the
closed-loop robot trajectories for the different VI-SLAM methods.
\textit{VINS} goes out of bounds for many runs. Focusing on the traces of
\textit{SVO} and \textit{GF}, it is clear that \textit{SVO} has a higher estimation variance across the
runs for a given sequence, while \textit{GF} trajectories are more closely clustered.
The properties hold irrespective of the IMU type.
Overall, \textit{GF} appears to be the strongst performer. As a modification of
\textit{ORB}, it seeks to reduce pose estimation latency while preserving the
beneficial properties of \textit{ORB}. The findings of this paper imply that
prioritizing accuracy while striving to achieve sufficiently small latencies
is an effective means to identifying a high performing VI-SLAM for
autonomous, mobile robot applications.  Some work is still needed to resolve
the outlier cases for \textit{GF}.

%



%
\section{Conclusion}
This paper investigated several stereo VI-SLAM methods to understand their
closed-loop trajectory tracking properties. The study was supported with a
simulated Gazebo environment 
shown to be representative of a specific set of benchmark conditions.
Analysis of the outcomes showed that both latency and drift play
important roles in achieving accurate trajectory tracking. A 
VI-SLAM system built upon ORB-SLAM, denoted by {\em GF}, provides 
the most accurate trajectory tracking outcomes, which is consistent 
with its open-loop performance.
Other methods were less consistent; SVO
has high performance in closed-loop but poor performance in
open-loop, and ORB vice-versa.
Future work will extend the benchmarking environments with 
additional mobile robots, rendering options, and visual environments or
settings.
Importantly, integration with actual collision-avoidance systems and the
impact of environmental obstacles on SLAM will improve the task-realism
of the benchmark.  





%
%
%

\balance 
\bibliographystyle{IEEEtran}
\bibliography{./full_references}

\end{document}